\documentclass{article}
\usepackage{spconf,amsmath,graphicx,hyperref, booktabs, comment}
\usepackage{siunitx}
\makeatletter
\def\title#1{\gdef\@title{\MakeUppercase{#1}}}
\def\name#1{\gdef\@name{\shortstack[c]{\itshape #1\\[0.5em]}}}
\makeatother

\title{Probing Spatial and Spectral Grounding in EEG Foundation Models}
\name{Kleanthis Avramidis$^1$,\quad Ivan Tashev$^{2}$,\quad Dimitra Emmanouilidou$^2$\thanks{This work was done during an internship at Microsoft Research.}
\address{$^1$ Signal Analysis and Interpretation Lab, University of Southern California, USA \\ $^2$ Microsoft Research, Redmond, WA, USA}
}

\begin{document}
\ninept
\maketitle
\begin{abstract}
EEG foundation models are advancing at an unprecedented pace, yet their evaluation is driven by downstream accuracy, which reveals little about whether their predictions rely on physiologically interpretable signal structure. We introduce a set of perturbation probes in two grounding axes: spatial grounding in channel identity and topographic locality, and spectral grounding in oscillatory content and waveform morphology. We apply these probes to five published EEG foundation models across motor-imagery and seizure recordings. Our experiments reveal differences in both representational sensitivity and downstream reliance. Electrode identity may be encoded without influencing downstream classification: permuting LaBraM's positional codes heavily displaced its embeddings yet reduced motor-imagery accuracy by only 1.2 points, while spectral details may be discriminable without producing specific responses to simulated clinical markers, as only REVE exceeded a spectral baseline on this task. These dissociations are hidden in common benchmarks, demonstrating how grounding probes can complement performance-based evaluation by revealing what EEG foundation models encode and what their predictions depend on.
\end{abstract}
\begin{keywords}
Electroencephalography, EEG foundation models, perturbation, spatial grounding, spectral grounding~\vspace{-2mm}
\end{keywords}
\section{Introduction}
\label{sec:intro}
\vspace{-1mm}

Electroencephalography (EEG) offers a non-invasive, low-cost measure of brain activity. EEG foundation models (EEG-FMs), trained on thousands of unlabeled recordings, have achieved state-of-the-art performance across domains ranging from motor imagery and sleep staging to emotion recognition and clinical event detection~\cite{cbramod,reve,kostas2021bendr}. However, task performance alone does not guarantee that a model captures  properties used in expert EEG interpretation. Such interpretation often relies on spatial context across electrodes and waveform characteristics when spectral information alone is insufficient.
For example, clinicians distinguish muscle contamination from epileptiform activity using the spatial distribution and morphology of a discharge, despite overlapping spectral properties and the dominance of muscle activity in scalp power~\cite{whitham2007}. More generally, electrode location carries physiologically meaningful information~\cite{nunez2006,kayser2015issues}, while clinically relevant EEG phenomena are often defined by waveform morphology~\cite{kural2020}. We refer to these complementary requirements as \emph{spatial} and \emph{spectral} grounding, where spectral grounding encompasses sensitivity not only to oscillatory power, but also to phase-dependent temporal structure and waveform morphology.

Such grounding may not be apparent from conventional downstream benchmarks: accuracy indicates how often an adapted model is correct, but not which signal properties support its predictions. This ambiguity is particularly relevant to EEG because recordings contain limited high-level semantic structure and are affected by environmental noise, device-specific artifacts, and between-subject variability. Consequently, competitive performance may be achieved partly through artifact-related shortcuts. Conversely, a representation may preserve meaningful spatial or spectro-temporal information that a downstream prediction head does not use. We therefore ask three separate questions: Can the architecture access a signal property? Is that property encoded in the learned representation? And do downstream predictions rely on it?

We investigate these distinctions through controlled perturbation-response experiments. To evaluate spatial grounding, we intervene on channel-to-location assignments by independently manipulating electrode identity, channel order, and learned positional encoding. Spectral probes test whether models distinguish recordings from surrogates that preserve power spectra while disrupting phase-dependent temporal structure, and whether their representations decode injected oscillations across canonical frequency bands. We further test whether representational sensitivity affects downstream predictions. We applied this framework to five EEG-FMs: BIOT~\cite{biot}, LaBraM~\cite{labram}, CBraMod~\cite{cbramod}, CSBrain~\cite{csbrain}, and REVE~\cite{reve}, and relate their observed dissociations to differences in model architecture and pre-training. Our contributions are as follows: 
\begin{itemize}
    \item We introduce spatial interventions that disentangle electrode identity and positional encoding, and characterize distance-dependent cross-channel influence across model depth.
    \item We introduce spectral interventions that evaluate sensitivity to phase-dependent structure beyond spectral power and to oscillatory activity across canonical frequency bands, calibrated against handcrafted, non-learned baselines.
    \item We show that architectural access to signal information, its preservation in learned representations, and its use in downstream predictions are distinct properties that cannot be inferred from performance benchmarks alone.
\end{itemize}

\textbf{Evaluation Studies}: Prior evaluation studies have found inconsistent gains over lighter supervised decoders: fine-tuned on memory and sleep benchmarks, EEG-FMs improve over these baselines by around 1\% while carrying millions of additional parameters~\cite{lee2025large}, whereas when the labels are scarce they do not outperform lighter decoders~\cite{yang2026eeg,aditya2026generalization}. Evaluation has accordingly moved past accuracy metrics. Benchmarking EEG-FMs on a suite of eight datasets yielded attribution maps that stay spatially stable under perturbation while the predictions degrade~\cite{vsirca2026beyond}. On the spectral side, a recent qualitative study injected synthetic oscillations and found that embeddings were dominated by the aperiodic power spectrum, with the oscillatory components increasingly underrepresented toward higher frequencies~\cite{kommineni2026aperiodic}. It remains unclear whether representations are grounded in interpretable EEG properties, such as electrode location or waveform structure beyond spectral power. 


\textbf{Foundation models for EEG.} EEG-FMs pretrain encoders on large unlabeled corpora and adapt them to diverse downstream tasks~\cite{li2024temporal}, differing in input tokenization and in how they encode electrode topology. We select 5 publicly released models that jointly cover the main design choices. \textbf{BIOT}~\cite{biot} tokenizes a fixed bipolar montage as STFT magnitudes indexed by learned slot tokens. \textbf{LaBraM}~\cite{labram} trains a quantization-based tokenizer whose targets are the Fourier amplitude and phase, then re-trains the encoder on masked code prediction. \textbf{CBraMod}~\cite{cbramod} adds a magnitude-spectrum pathway and applies a depthwise convolution over the channel axis, consuming order but not electrode identity. \textbf{CSBrain}~\cite{csbrain} shares that magnitude pathway while grouping electrodes into anatomical regions, thus attending over coarse topography. Finally, \textbf{REVE}~\cite{reve} embeds raw temporal patches and conditions attention directly on continuous 3D electrode coordinates. The models further differ in size, ranging from $3.2$M parameters for BIOT to $69$M for REVE, and are trained on corpora ranging from a few   to thousands of hours.\vspace{-2mm}


\section{Data and Pre-processing}
\label{ref:data}
\vspace{-1mm}

We used two datasets that differ in montage, environment, and task. \textbf{PhysioNet-MI}~\cite{schalk2004bci2000,pollard_physionet_2026} is a 4-class motor imagery corpus with 109 participants at 64 channels, of which we only probe the 20 participants (1,805 samples) that define the class-balanced test set in prior studies~\cite{cbramod} to avoid leakage. PhysioNet-MI is pre-processed in line with previous studies. Recordings are average-referenced, high-pass filtered at 0.3\,Hz, notch filtered at 60\,Hz, resampled to 200\,Hz, and epoched from cue onset to 4\,s without baseline correction.

\textbf{Siena Scalp}~\cite{detti2020eeg,pollard_physionet_2026} provides a 2-class clinical setting of seizure versus non-seizure windows, recorded from 14 participants with 29 electrodes. Recordings are band-passed at 0.1–75\,Hz, notch filtered at 50\,Hz, and resampled to 200\,Hz. Our probing uses a subject-stratified balanced subset of 374 windows of 10\,s from the 4 held-out subjects from~\cite{cbramod,csbrain}. Across both corpora, the cached waveforms remain unnormalized so that each encoder applies its own input scaling. Electrode coordinates are drawn from the standard 10–05 montage, mean-centered, and divided by the largest norm.

All five EEG-FMs are kept frozen throughout our protocol and only linear readouts are probed. Unless noted, $f{(\mathbf{X})}$ denotes penultimate encoder activations: tokens are pooled within output channel and the channel vectors concatenated. BIOT is probed over the 16 bipolar derivations, each assigned the midpoint of its electrode pair.\vspace{-2mm}

\section{Spatial Grounding}
\label{sec:spatial}
\vspace{-1mm}

Each EEG channel reflects a position-dependent mixture of underlying neural sources~\cite{nunez2006}. We therefore test whether pretrained model representations retain channel identity and topographic locality, and whether  sensitivity to  the spatial location may influence downstream predictions. EEG-FMs encode this information differently: BIOT assigns learned channel-slot tokens, LaBraM indexes electrode names, CSBrain groups electrodes into anatomical regions, while REVE explicitly encodes positional embeddings of 3D coordinates on the scalp. CBraMod on the other hand operates on a fixed order without explicit electrode identities. We design three probes to distinguish these mechanisms and test their downstream influence.\vspace{-2mm}

\subsection{Spatial Probes}

\textbf{SPA1: Sensitivity to spatial encoding.} Let $f_\pi(\mathbf{X})$ denote the channel-mean pooled embedding of a sample $\mathbf{X}$ whose channels carry the positional labels $\pi$. We derange the labels to $\pi'$ while keeping waveforms fixed, and compute the relative displacement
\[
  S_{\mathrm{id}} \;=\; \big\lVert f_{\pi'}(\mathbf{X}) - f_{\pi}(\mathbf{X}) \big\rVert_2 \,\big/\, \big\lVert f_{\pi}(\mathbf{X}) \big\rVert_2 \,.
\]
$S_{\mathrm{id}} = 0$ indicates that the representation is invariant to the assigned channel positions, while larger values indicate stronger reliance on spatial encoding. Since CBraMod infers spatial structure from the order of channels in the input, we compute an analogous displacement $S_{\mathrm{ord}}$ after permuting the input order of the channels. All other models were invariant to this permutation. Finally, for all architectures we compute $S_{\mathrm{sig}}$, the displacement after permuting the waveforms across channels while keeping each channel ID fixed, which breaks the signal-to-location correspondence.

\textbf{SPA2: Locality of cross-channel influence.} We test whether cross-channel influence retains the spatial locality expected of EEG. Each input channel $i$ is perturbed in turn with white noise $\epsilon_i$ at $10$\,dB SNR relative to its power within the sample. Output tokens are flattened per channel, so the influence of $i$ on output channel $j$ is
\begin{equation}
d_{ij} = \lVert f_j(X + \epsilon_i) - f_j(X)\rVert_F \,/\, \lVert f_j(X)\rVert_F .
\label{eq:influence}
\end{equation}
where $f_j$ denotes the channel-level representation. We summarize the resulting influence matrix by the fraction of displacement occurring outside the perturbed channel $\sum_{j\neq i} d_{ij}/\sum_j d_{ij}$ and by the Spearman correlation between off-channel influence $d_{ij}$ ($j\neq i$) and inter-electrode distance; the perturbed channel itself is excluded. The probe is applied after each transformer block.

\textbf{SPA3: Spatial bias in model predictions.} Finally, we test whether spatial sensitivity in the representation affects model predictions in a downstream task. We fit a linear head on clean features from PhysioNet-MI and, at test time, we permute the positional information while the input remains unchanged. Specifically, we permute the coordinates supplied to REVE, LaBraM's electrode indices, BIOT's channel-slot tokens, and CSBrain's region labels (thus excluding CBraMod). We report both the latent displacement $\lVert\delta f\rVert/\lVert f\rVert$ and the change in balanced accuracy. PhysioNet-MI is used because its labels depend on lateralized activity.

\subsection{Experimental setup}

SPA1 was evaluated on PhysioNet-MI and Siena using 10 derangements per intervention. For each model and dataset, 64-component PCA was fitted to  clean embeddings and applied unchanged to  perturbed representations, controlling for differences in representation dimensionality. Error bars denote standard deviations across derangements. SPA2 was evaluated on PhysioNet-MI at every encoder block through the penultimate layer. SPA3 used 5-fold subject-disjoint cross-validation. Within each fold,  PCA fitted to the clean  training embeddings and applied to all the clean and positionally permuted embeddings of held-out participants. Results are averaged over three positional permutations per fold.

\section{Spectral Grounding}
\label{sec:spectral}

EEG-FMs are commonly pretrained through masked reconstruction in either the temporal or spectral domain~\cite{labram, reve, liu2026eeg}, encouraging them to follow the dominant $1/f$ aperiodic spectrum~\cite{kommineni2026aperiodic}. Clinically relevant information, however, also depends on oscillatory activity and waveform structure not determined by spectral power alone~\cite{donoghue2020parameterizing}. The five considered models impose different spectral biases: BIOT consumes only STFT magnitudes, LaBraM trains its tokenizer against Fourier amplitude and phase, CBraMod and CSBrain add magnitude-spectrum projections to their embeddings, while REVE operates solely on temporal patches. We design three probes for information beyond phase-free power, oscillatory sensitivity, and decision-level specificity to clinical morphology.

\subsection{Spectral Probes}
\label{sec:spe}

\textbf{SPE1: Information beyond spectral power.} For each recording, we generate an iterative amplitude-adjusted Fourier-transform (IAAFT) surrogate~\cite{schreiber1996improved} that preserves each channel's amplitude distribution and approximately its power spectrum. Phases are randomized with the same random phase sequence across channels, approximately preserving inter-channel phase differences~\cite{prichard1994generating}. A linear head is then trained to distinguish original from surrogate embeddings; above-chance accuracy indicates that the representation encodes waveform structure beyond amplitude distribution and power spectrum. Because IAAFT leaves a small spectral residual and removes nonstationary power fluctuations, we benchmark each encoder against phase-free spectral power on the same task.

\textbf{SPE2: Frequency-specific oscillatory sensitivity.} We inject a narrowband oscillation into one of four canonical bands, namely delta, theta, alpha, or beta (ranges are given in Figure~\ref{fig:spe2}), and train a four-way linear head to identify the injected band from the latent displacement. Every channel receives an independent carrier produced by passing white Gaussian noise through a fourth-order Butterworth filter at the band boundaries. Injection power is measured against the recording’s original power in that same band, preventing the aperiodic spectrum from disproportionately weighting frequencies. The latent displacement was also unit-normalized so that band identity cannot be inferred from displacement magnitude alone.

\textbf{SPE3: Sensitivity to clinical morphology.} Finally, we test whether spectral sensitivity carries over to clinical predictions. A linear seizure classifier is fit on clean Siena Scalp representations, and at test time we inject into held-out non-seizure segments (at $0$\,dB) either a rhythmic ictal discharge, a spike-wave pattern, or power-matched pink noise control. The ictal marker is a chirp decreasing from $10$ to $3$\,Hz with increasing amplitude; the spike-wave marker repeats at $3$\,Hz, each event being a Gaussian-derivative spike with extrema $70$\,ms apart~\cite{kural2020}, followed after $100$\,ms by a $100$\,ms Gaussian slow wave. Each waveform is injected identically into a contiguous 20--40\% channel subset formed by a seed electrode and its nearest neighbors. We report specificity as the mean change in $P(\mathrm{seizure})$ for the two epileptiform patterns minus that for pink noise.\vspace{-3mm}

\begin{figure}
    \centering
    \includegraphics[width=0.98\linewidth]{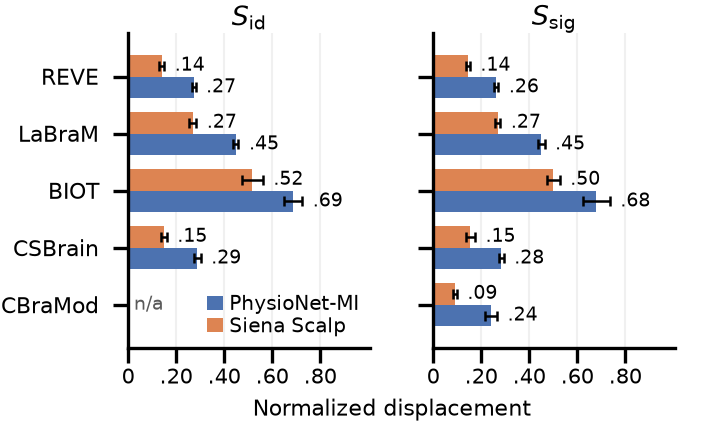}
    \vspace{-3mm}
    \caption{\textbf{SPA1 --- Sensitivity to spatial encoding.} Normalized latent displacement after permuting electrode labels while holding the waveforms fixed ($S_{\mathrm{id}}$; left) and after breaking the correspondence between signals and their location ($S_{\mathrm{sig}}$; right).}
    \label{fig:spa1}
    \vspace{-5mm}
\end{figure}

\subsection{Experimental setup}

SPE1 used subject-disjoint 5-fold cross-validation for PhysioNet-MI and leave-one-subject-out for Siena. Within each fold, PCA with 64 components is fitted to clean training data and applied unchanged to the corresponding training surrogates and all held-out samples. Logistic regression is then fitted using the clean and surrogate training examples. Uncertainty is estimated across participants. SPE2 uses the same PCA and cross-validation on PhysioNet-MI. A four-way linear classifier is fitted to the unit-normalized latent displacements and evaluated on held-out participants. Results are averaged across 3 carrier realizations per band and injection level, which varied from $-20$ to $+5$\,dB relative to the original band power. SPE3 uses the same cross-validation on Siena. Within each fold, the seizure classifier is fitted on all samples and interventions are applied only to the non-seizure windows of the held-out participant.\vspace{-2mm}

\begin{figure}
    \centering
    \includegraphics[width=0.94\linewidth]{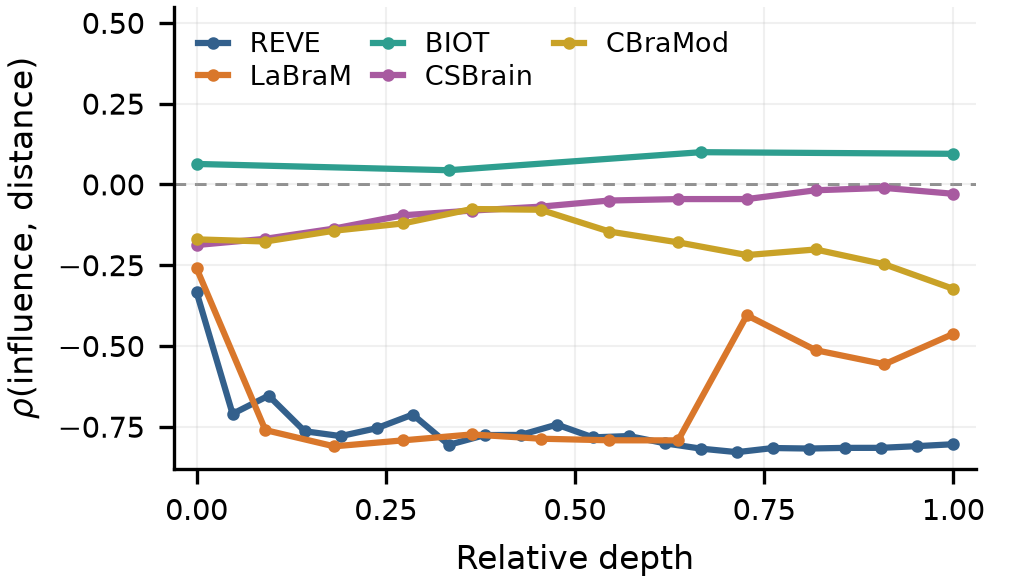}
    \vspace{-4mm}
    \caption{\textbf{SPA2 --- Topographic structure across depth.} Spearman correlation between off-channel influence and inter-electrode distance. Negative values indicate stronger influence on nearby than distant channels. Results are shown on PhysioNet-MI.}
    \label{fig:spa2}
    \vspace{-2mm}
\end{figure}

\begin{table}[t]
\centering
\setlength{\tabcolsep}{4.8pt}
\begin{tabular}{@{}lcccc@{}}
\toprule
\textbf{Model} & \textbf{Clean} & \textbf{Permuted} & $\Delta$ & $\lVert\delta f\rVert/\lVert f\rVert$ \\
\midrule
REVE & 0.455$_{\pm0.006}$ & 0.340$_{\pm0.012}$ & \textbf{+0.115}$_{\pm0.013}$ & 0.714 \\
LaBraM & 0.354$_{\pm0.004}$ & 0.343$_{\pm0.011}$ & +0.012$_{\pm0.010}$ & 0.805 \\
BIOT & 0.276$_{\pm0.012}$ & 0.275$_{\pm0.004}$ & +0.001$_{\pm0.009}$ & 0.589 \\
CSBrain & 0.440$_{\pm0.008}$ & 0.337$_{\pm0.004}$ & +0.104$_{\pm0.008}$ & 0.885 \\
\bottomrule
\end{tabular}
\vspace{-2mm}
\caption{\textbf{SPA3} --- Balanced accuracy before and after permuting the learned positional code at test time on PhysioNet-MI, with the input waveform unchanged and the readout frozen. The last column reports the induced latent change. Random chance is 25\%.}
\label{tab:spa3}
\vspace{-4mm}
\end{table}

\section{Results and Discussion}
\label{sec:results}
\vspace{-1mm}

\subsection{Spatial grounding}

The spatial probes reveal distinct mechanisms for representing channel location. EEG-FMs with explicit positional encoding were largely invariant to channel order, whereas CBraMod was built directly to process it. Among identity-aware models, BIOT's learned slot tokens produced the largest identity displacement (0.687 on PhysioNet-MI, 0.517 on Siena Scalp) whereas REVE's 3D coordinates produced the smallest (0.272 on PhysioNet-MI, 0.140 on Siena Scalp). Breaking the signal-to-location correspondence produced a similar $S_{\mathrm{sig}}$ ordering on both datasets, reflecting consistent model-specific sensitivity across domains (Fig.~\ref{fig:spa1}), with lower Siena values consistent with its fewer channels.

SPA2 also found spatial locality to be strongly depth-dependent, as shown in Fig.~\ref{fig:spa2}. Perturbations in REVE and LaBraM affected nearby electrodes more than distant ones ($\rho=$ -0.829 and -0.811, respectively), with the strongest topographic organization appearing in intermediate layers for the latter. CSBrain and CBraMod showed little distance structure, while BIOT was approximately flat. In contrast, the fraction of displacement outside the perturbed channel converged for all models to 0.720–0.868, therefore model representations become broadly mixed, yet that does not systematically preserve or discard their spatial biases. The task-level intervention on PhysioNet-MI further separated representational sensitivity from functional reliance (Table~\ref{tab:spa3}). Permuting positional codes substantially altered the representations of REVE, LaBraM, and CSBrain, but LaBraM did not show a corresponding loss in motor-imagery accuracy. BIOT's large SPA1 identity displacement could not be assessed at the decision level, as its readout performed near chance.\vspace{-2mm}

\subsection{Spectral grounding}

\begin{figure}
    \centering
    \includegraphics[width=0.94\linewidth]{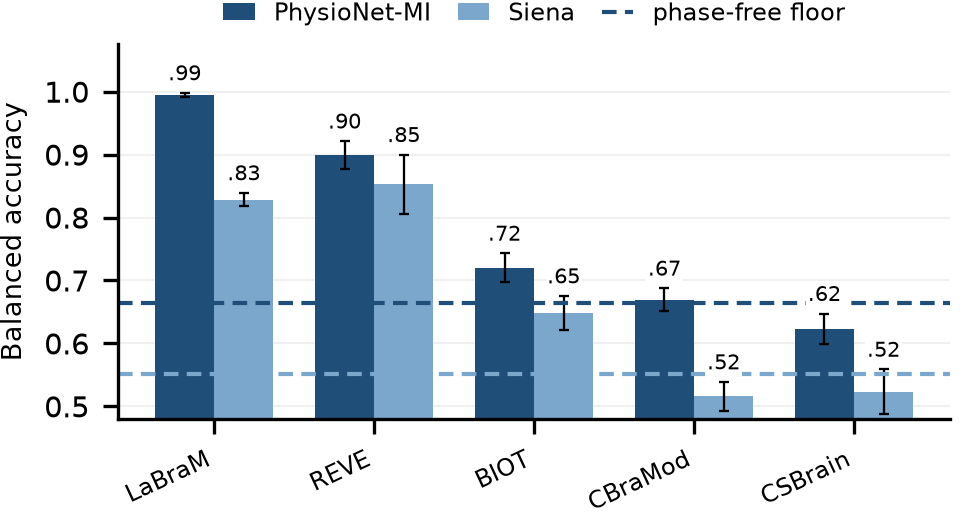}
    \vspace{-4mm}
    \caption{\textbf{SPE1 --- Information beyond spectral power.} Balanced accuracy of linear probes distinguishing clean recordings from IAAFT~\cite{schreiber1996improved} surrogates on PhysioNet-MI and Siena. Error bars indicate SD from participant-clustered bootstrapping at 200 resamples.}
    \label{fig:spe1}
    \vspace{-5mm}
\end{figure}

\begin{figure}
    \centering
    \includegraphics[width=0.96\linewidth]{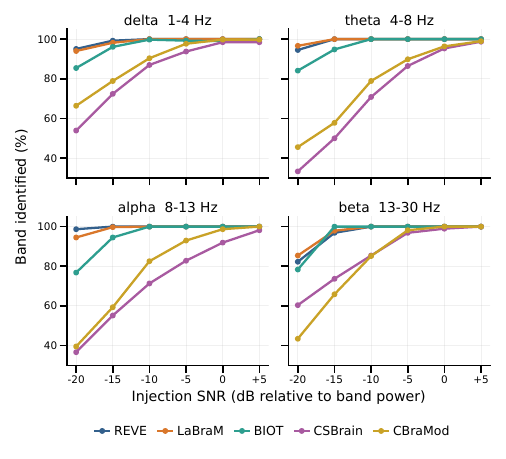}
    \vspace{-6mm}
    \caption{\textbf{SPE2 --- Sensitivity to narrowband oscillations.} Four-way accuracy for identifying the injected delta, theta, alpha, or beta band from the unit-normalized latent displacement, shown against injection power relative to the original band power.}
    \label{fig:spe2}
\end{figure}

SPE1 showed that most EEG-FMs only partially retain structure beyond spectral power on the two corpora (Fig.~\ref{fig:spe1}). REVE and LaBraM in particular clearly distinguished EEG from the phase-randomized surrogate, whereas BIOT was marginally better than a phase-free control derived from spectral power alone. CSBrain and CBraMod scored at or below that baseline, consistent with representations dominated by their magnitude-spectrum pathways. Sensitivity to band injection in SPE2 followed a similar pattern, shown in Fig.~\ref{fig:spe2}. REVE and LaBraM were able to identify the injected frequency band at substantially lower perturbation levels than the other models, while CSBrain and CBraMod collapsed under weaker injections. Sensitivity was also frequency-dependent: CSBrain and CBraMod degraded most for weaker theta and alpha injections, while performance recovered in the beta band.

Finally, according to SPE3, changes in seizure probability in the Siena Scalp test set were less specific than the embedding-level results suggest (Table~\ref{tab:spe3}). All encoders increased their seizure-class probability following at least one epileptiform intervention, but several also responded strongly to matched-power pink noise. REVE was the only model for which noise moved predictions consistently away from the seizure class and whose epileptiform specificity exceeded the log-PSD control. LaBraM responded to the perturbations almost as strongly as to noise, illustrating that spectral sensitivity alone does not establish clinically meaningful waveform selectivity.\vspace{-2mm}

\subsection{Representations do not predict decision reliance}

Together, our spatial and spectral probes reveal model-specific profiles that are not recoverable from architecture descriptions or downstream rankings alone. BIOT's slot tokens provided the strongest channel identity displacement but these changes carried no measurable decision-level reliance in this task. CSBrain relied on its region code for motor-imagery decisions despite showing little topographic locality, whereas CBraMod showed little sensitivity to either spatial or spectral conditioning. REVE’s explicit 3D electrode coordinates were associated with the clearest positional and morphological decision effects in our probes, even though it showed the smallest identity displacement in SPA1. Likewise, LaBraM’s robust surrogate and oscillation decoding did not translate into specificity to clinical morphology. We deduce that evaluation protocols for EEG foundation models should test not only whether a representation contains discriminable or purely physiological structure, but also whether downstream decisions are sensitive to that structure.\vspace{-3mm}

\begin{table}[t]
\centering
\footnotesize
\setlength{\tabcolsep}{3.5pt}
\begin{tabular}{@{}lcccc@{}}
\toprule
\textbf{Model} & \textbf{Ictal rhythmic} & \textbf{Spike wave} & \textbf{Pink noise} & \textbf{Specificity} \\
\midrule
REVE & +0.113$_{\pm0.027}$ & +0.097$_{\pm0.029}$ & -0.053$_{\pm0.011}$ & \textbf{+0.158}$_{\pm0.018}$ \\
LaBraM & +0.131$_{\pm0.027}$ & +0.055$_{\pm0.015}$ & +0.115$_{\pm0.025}$ & -0.022$_{\pm0.039}$ \\
BIOT & +0.069$_{\pm0.059}$ & +0.053$_{\pm0.045}$ & -0.014$_{\pm0.032}$ & +0.076$_{\pm0.071}$ \\
CSBrain & +0.114$_{\pm0.035}$ & +0.053$_{\pm0.021}$ & +0.044$_{\pm0.005}$ & +0.040$_{\pm0.013}$ \\
CBraMod & +0.036$_{\pm0.008}$ & +0.080$_{\pm0.014}$ & +0.040$_{\pm0.008}$ & +0.019$_{\pm0.006}$ \\
\midrule
log-PSD & +0.117$_{\pm0.023}$ & +0.103$_{\pm0.020}$ & +0.083$_{\pm0.023}$ & +0.027$_{\pm0.025}$ \\
\bottomrule
\end{tabular}
\vspace{-2mm}
\caption{\textbf{SPE3} --- Change in $P(\mathrm{seizure})$ after a biomarker is injected into held-out non-seizure segments of Siena Scalp at $0$\,dB, with the readout fitted on clean folds held frozen. Specificity is the mean epileptiform response minus the noise control response. Values are participant-level means $\pm$ standard error of the mean (SEM).}
\label{tab:spe3}
\vspace{-4mm}
\end{table}

\section{Conclusion}
\label{sec:conclusion}
\vspace{-2mm}

We introduced a set of perturbation probes to evaluate whether EEG foundation models encode and use physiologically meaningful spatial and spectral structure. Across five models, the resulting profiles were not determined by architecture or captured by downstream accuracy. Channel identity could be strongly encoded without topographic locality (BIOT) or without influencing classification (LaBraM), or be relied upon only at a coarse regional level (CSBrain), while oscillatory sensitivity did not imply selectivity for pathological morphology. What a representation encodes therefore need not govern its predictions. That said, our analysis is constrained by the limited number of Siena Scalp participants and by using only linear readouts on frozen encoders, which capture only linearly accessible information and may not reflect fully fine-tuned models. Broader ecological and clinical validation is therefore needed.

\section{Compliance with Ethical Standards}

This study is a secondary analysis of publicly available, de-identified human EEG data. No new human-participant data were collected. Ethical approval and informed consent for the original data collections are documented in the corresponding dataset publications.

\bibliographystyle{IEEEbib}
\bibliography{refs}

\end{document}